%% file: main.tex
\documentclass[conference]{IEEEtran}
\usepackage{times}

\usepackage[numbers]{natbib}
\usepackage{multicol}
\usepackage[bookmarks=true]{hyperref}
\usepackage{subfig}
\usepackage{graphicx}
\usepackage{xcolor}
\usepackage{colortbl}
\usepackage{multirow}
\usepackage{tcolorbox}
\usepackage{booktabs} 
\usepackage{wrapfig}

\input{notion_marcos}

\begin{document}

\title{Reimagination with Test-time Observation Interventions: Distractor-Robust World Model Predictions for Visual Model Predictive Control}

% You will get a Paper-ID when submitting a pdf file to the conference system
\author{
    {Yuxin Chen}$^{\ast,1}$, {Jianglan Wei}$^{\ast,1}$, {Chenfeng Xu}$^{1}$, {Boyi Li}$^1$, {Masayoshi Tomizuka}$^1$, {Andrea Bajcsy}$^2$, {Ran Tian}$^{1}$ \\
    $^{\ast}$Denotes equal contribution \\
    $^1$\textit{University of California, Berkeley} \quad $^2$\textit{Carnegie Mellon University}
}
% \author{\authorblockN{Michael Shell}
% \authorblockA{School of Electrical and\\Computer Engineering\\
% Georgia Institute of Technology\\
% Atlanta, Georgia 30332--0250\\
% Email: mshell@ece.gatech.edu}
% \and
% \authorblockN{Homer Simpson}
% \authorblockA{Twentieth Century Fox\\
% Email: homer@thesimpsons.com}
% \and
% \authorblockN{James Kirk\\ and Montgomery Scott}
% \authorblockA{Starfleet Academy\\
% San Francisco, California 96678-2391\\
% Telephone: (800) 555--1212\\

% avoiding spaces at the end of the author lines is not a problem with
% conference papers because we don't use \thanks or \IEEEmembership

%Homer Simpson\authorrefmark{2},
%Montgomery Scott\authorrefmark{3} and
%Eldon Tyrell\authorrefmark{4}}
%Telephone: (800) 555--1212, Fax: (888) 555--1212}
%\authorblockA{\authorrefmark{4}Tyrell Inc., 123 Replicant Street, Los Angeles, California 90210--4321}}

\maketitle

\begin{abstract}
World models enable robots to ``imagine” future observations given current observations and planned actions, and have been increasingly adopted as generalized dynamics models to facilitate robot learning.
Despite their promise, these models remain brittle when encountering novel visual distractors such as objects and background elements rarely seen during training. Specifically, novel distractors can corrupt action outcome predictions, causing downstream failures when robots rely on the world model imaginations for planning or action verification. 
In this work, we propose Reimagination with Observation Intervention (\textbf{ReOI}), a simple yet effective test-time strategy that enables world models to predict more reliable action outcomes in open-world scenarios where novel and unanticipated visual distractors are inevitable.
Given the current robot observation, ReOI first detects visual distractors by identifying which elements of the scene degrade in physically implausible ways during world model prediction. Then, it modifies the current observation to remove these distractors and bring the observation closer to the training distribution. Finally, ReOI ``reimagines” future outcomes with the modified observation and reintroduces the distractors post-hoc to preserve visual consistency for downstream planning and verification.
We validate our approach on a suite of robotic manipulation tasks in the context of action verification, where the verifier needs to select desired action plans based on predictions from a world model.
Our results show that \textbf{ReOI} is robust to both in-distribution and out-of-distribution visual distractors. Notably, it improves task success rates by up to \textbf{3×} in the presence of novel distractors, significantly outperforming action verification that relies on world model predictions without imagination interventions.
\end{abstract}

\IEEEpeerreviewmaketitle

%==================================================

\input{sections/intro}

\input{sections/related_works}

\input{sections/preliminaries}

\input{sections/method}

\input{sections/experiment_setup}

\input{sections/results}

\input{sections/conclusion}

% \input{sections/limitation}

%% Use plainnat to work nicely with natbib. 

\bibliographystyle{plainnat}
\bibliography{references}

\clearpage
\appendix

\input{sections/appendix}
\end{document}

%% file: notion_marcos.tex
\usepackage{amsmath,amssymb,stmaryrd,mathtools}
\usepackage{xcolor}
\usepackage{xspace}
\usepackage{bbm}
\usepackage{graphicx}
\usepackage{subfig} 
\usepackage{tcolorbox}

\definecolor{dark_red}{RGB}{122, 0, 0}
\definecolor{coral}{RGB}{255, 119, 94}
\definecolor{pink_orange}{RGB}{255, 72, 126}
\definecolor{vibrant_pink}{RGB}{255, 0, 104}
\definecolor{pink_pink}{RGB}{255, 37, 153}
\definecolor{wine}{RGB}{204, 0, 102}

\definecolor{light_orange}{RGB}{255, 198, 107}
\definecolor{orange(sae/ece)}{rgb}{1.0, 0.49, 0.0}
\definecolor{dark_orange}{RGB}{216,92,0}

\definecolor{org-purp-0}{RGB}{165, 76, 0}
\definecolor{org-purp-1}{RGB}{250, 130, 28}
\definecolor{org-purp-2}{RGB}{226, 89, 68}
\definecolor{org-purp-3}{RGB}{206, 92, 124}
\definecolor{org-purp-4}{RGB}{116, 80, 146}
\definecolor{org-purp-5}{RGB}{110, 78, 157}

\definecolor{teal(sae/ece)}{rgb}{0, 0.47, 0.52}
\definecolor{aqua}{RGB}{52,172,139}
\definecolor{dark_aqua}{RGB}{35,115,93}
\definecolor{dark_green}{RGB}{0, 92, 34}

\definecolor{grape}{RGB}{112,48,160}
\definecolor{purple}{rgb}{0.74, 0.65, 1.0}
\definecolor{dark_purple}{rgb}{0.58, 0.0, 0.82}
\definecolor{periwinkle}{RGB}{191, 140, 230}

\definecolor{light_gray}{rgb}{0.9, 0.9, 0.9}
\definecolor{medium_gray}{rgb}{0.6, 0.6, 0.6} 
\definecolor{dark_gray}{rgb}{0.2, 0.2, 0.2} 

\definecolor{sky_blue}{RGB}{37, 166, 213}
\definecolor{light_blue}{rgb}{0.33, 0.80, 1}
\definecolor{dark_blue}{rgb}{0.098, 0.239, 0.52}
\definecolor{ocean}{RGB}{13, 121, 202}
\definecolor{light_ocean}{RGB}{18, 178, 235}
\definecolor{dark_ocean}{RGB}{10, 89, 148}
\definecolor{vibrant_blue}{RGB}{14, 120, 255}

\definecolor{dark_brown}{rgb}{0.3255, 0.004, 0.001}

\newcounter{qnum}
\newcounter{tnum}
\newcommand{\ours}{\textcolor{org-purp-1}{\textbf{ReOI}}\xspace}
\newcommand{\dinowm}{\textcolor{medium_gray}{\textbf{DINO-WM}}\xspace}

\newcommand{\trust}{\textcolor{sky_blue}{\textbf{TrustRegion}}\xspace}

\newcommand{\obs}{o}

\newcommand{\obstraj}{\mathbf{o}}

\newcommand{\dyn}{f_\phi}

\newcommand{\lang}{\ell}

\newcommand{\action}{a}
\newcommand{\acttraj}{\mathbf{\action}}

\newcommand{\thor}{T}

\usepackage{amsmath}

%% file: sections/intro.tex
\section{Introduction}

World models \cite{ha2018recurrent,zhou2024dino} enable robots to predict action-conditioned future evolutions of their environments given current observations and planned actions. They have emerged as powerful generalized dynamics models and are increasingly used in model-based policy learning \cite{qi2025strengthening, dedieu2025improving, wu2023daydreamer} as well as in deployment-time action plan verification \cite{gaoflip, wu2025foresight}.
However, despite their promising potential, current world models in robotics are brittle against visual distractors such as task-irrelevant objects or background elements rarely seen during training.
These novel distractors can corrupt the model, leading to hallucinated imaginations that ultimately compromise downstream action plan verification and selection.
Notably, this brittleness persists even when models are trained on large, visually diverse datasets sourced beyond robotics \cite{agarwal2025cosmos,zhenlearning}.

To illustrate this challenge, consider a robot assistant tasked with meal preparation, picking up prepped ingredients and placing them into a cooking pan.  
The robot uses a world model, trained on demonstrations from similarly structured kitchens, to verify candidate action plans proposed by a pre-trained imitation policy. Specifically, the robot uses the world model to imagine the future outcomes of each action plan and evaluates them using a visual reward function to identify the best action plan (i.e., policy verification).
However, in open-world deployment, novel visual distractors are inevitable. For example, a user might place an unfamiliar high-pressure pot between the cutting board and the pan, or leave an Amazon package box in the background as illustrated in Figure~\ref{fig:front-figure}.
As the world model rolls out candidate action plans, these unfamiliar distractors often degrade in visually implausible ways, becoming distorted, disappearing, or warping unnaturally across predicted frames.
These artifacts cause the model to misrepresent distractors that are critical for ensuring safety and hallucinate incorrect robot behaviors.
As a result, the downstream plan verifier failed to detect that a planned motion that would collide with the pot, ultimately leading to deployment-time failures (Figure~\ref{fig:front-figure}, top).

\input{sections/front_figure}

These failure cases expose a limitation in current world model–based robot policy verification pipelines and motivate the need for strategies that can mitigate the effects of novel distractors. 
Our key insight is that relying solely on training-time solutions is insufficient: the diversity and unpredictability of distractors, especially in open-world settings, make it impractical to fully capture them, even with large-scale datasets.
To this end, we propose Reimagination with Observation Intervention, \ours, a \textit{test-time} and \textit{plug-in} strategy that mitigates the impact of novel visual distractors on world model-based robot policy verification. 
Our key idea is to first identify and inpaint novel distractors from the current observation to bring it closer to the world model's training distribution, then reimagine future action outcomes using the modified input, and finally reintroduce the distractors post-hoc at the pixel level to preserve visual consistency for downstream planning and verification.

We validate our approach on robotic manipulation tasks in the context of action plan verification, where a verifier needs to select action plans based on visual outcome predictions from a world model.
Our approach demonstrates strong robustness to both in-distribution and unfamiliar visual distractors. Notably, our method improves task success rates by up to \textbf{3x} when encountering novel distractors compared to standard policy verification that directly relies on unmodified world model predictions.

\begin{tcolorbox}[colback=orange!6!white, colframe=white]
\textbf{Contribution}. In this work, we raise and tackle the challenge of mitigating the effects of novel visual distractors on world model-based robot policy verification.
We illustrate concrete failure cases induced by such distractors and introduce Reimagination with Observation Intervention (\ours), a \textit{test-time}, \textit{plug-in} strategy that enables world models to produce more reliable action outcome predictions in open-world scenarios where novel and unanticipated distractors are inevitable.
To the best of our knowledge, this is the first work that leverages test-time observation intervention to address the problem of novel visual distractors in world model–based robot policy verification and selection.
\end{tcolorbox}

%
% While this observation intervention strategy shifts the input closer to the world model’s training distribution—enabling more stable and accurate predictions by effectively pretending the novel distractors are not present—the post-generation reintroduction of distractors can lead to physically unrealistic interactions, such as the robot visually passing through them. However, this does not compromise action plan verification. Since the robot must avoid any physical contact with such distractors to ensure safety, any trajectory exhibiting these violations is automatically rejected during the verification process.

% We also compare our approach against an test-time out-of-distribution (OOD) detection method that rejects unconfident world model rollouts in the context of policy verification. 
% %Our method achieves performance comparable to the training-time strategy when dealing with in-distribution distractors, while significantly outperforming it in the presence of unfamiliar distractors. 
% %
% %Additionally
% Compared to the OOD detection baseline, our approach is much less conservative, resulting in higher task success rates and fewer false rejections of safe action plans. 
% %
% These results underscore the value of inference-time adaptation for handling visual distractors, particularly in open-world scenarios where novel and unanticipated distractors are inevitable.

%% file: sections/front_figure.tex
\begin{figure*}[t]
  %\vspace{-0.8cm}
  \centering
\includegraphics[width=0.8\textwidth]{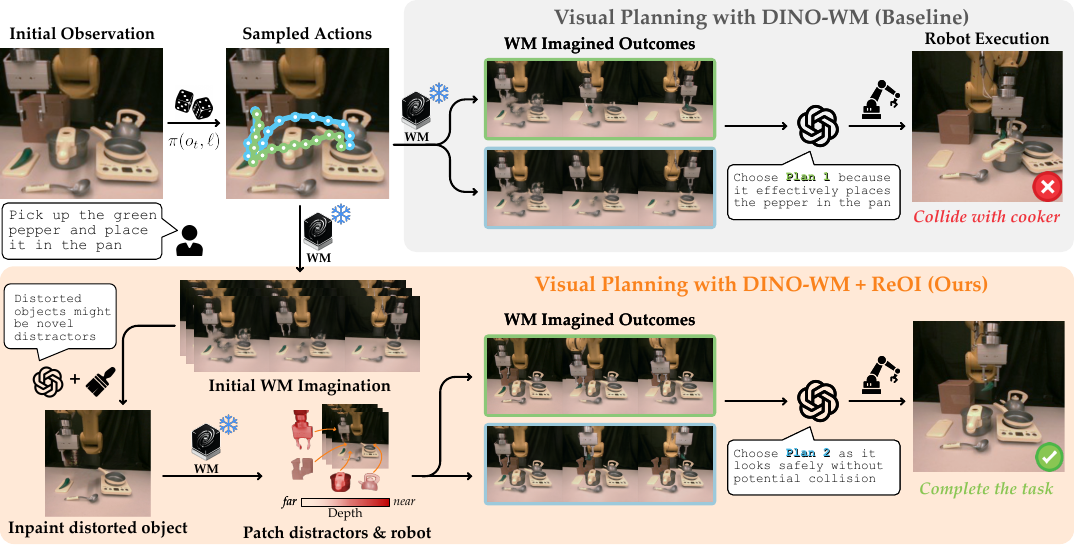}
  \caption{\textbf{Reimagination with observation intervention for distractor-robust world model-based robot planning}. A robot uses a world model (WM) to verify and select action plans proposed by a pre-trained imitation policy. The world model has never encountered the box, cooker, or teapot during training. \textbf{Top (baseline)}: Novel visual distractors become distorted across predicted observations, causing the model to hallucinate incorrect robot behaviors and erase a critical obstacle. This leads the verifier to select an unsafe action plan. \textbf{Bottom (ours)}: Test-time observation intervention enables the world model to generate predictions better aligned with reality, allowing the verifier to recognize potential collisions and correctly select a safe action plan.}
  \vspace{-0.6cm}
  \label{fig:front-figure}
\end{figure*}

%% file: sections/related_works.tex
\section{Related Works}
\vspace{-0.2cm}

\textbf{Mitigating visual distractors in world model learning.} Learning accurate world models in visually cluttered environments remains a challenge in robotics.
Most existing approaches mitigate visual distractors through training-time strategies: either by leveraging privileged reward signals to learn task-relevant representations \cite{zhang2020learning, fu2021learning, wang2022denoised, zhu2023repo}, or by training separate network branches to model task-relevant and task-irrelevant components, using only the task-relevant branch during downstream policy planning or verification \cite{wan2023semail, wang2024ad3, huang2024leveraging}.
However, these methods primarily target in-distribution visual distractors and are tightly coupled to the training-time task context. When the task context shifts at test time, causing previously irrelevant distractors to become safety-critical or introducing novel distractors, their performance often degrades and even fails entirely~\cite{ho2025worldmodel}. 
%
% In contrast, we argue that novel visual distractors are inevitable in open-world deployments—and may even be safety-critical.
\textit{Rather than relying on task-specific supervision to suppress distractors during world model training and assuming the same context will hold at deployment, we propose a test-time observation intervention strategy to mitigate novel visual distractors' impact on world model–based robot policy verification.}

\textbf{Observation intervention for improving robot policy robustness.} 
Modifying robot observations by masking out the background or task-irrelevant objects has proven effective in improving the robustness of \textit{imitation policies} to visual distractors. Prior works have explored using separately trained models \cite{miyashita2023roso,riedmiller2023less}, VLMs \cite{yang2025transferring}, or conformal prediction techniques \cite{hancock2024run} to identify and inpaint visual distractors to improve policy performance.
\textit{Different from these works that focus on modifying observation to improve model-free imitation policy robustness, our work focuses on test-time observation intervention for improving the reliability of model-based visual planning in the presence of novel visual distractors.}

% \textbf{World-model for robot learning.}

%% file: sections/preliminaries.tex
\section{Preliminaries}

\textbf{Sampling-based visual model predictive control}.
We formulate the robot planning problem as a sampling-based visual model predictive control problem where the robot uses a predictive world model to forward-simulate multiple action-conditioned futures and evaluates them using a reward function to select the best plan for execution. Mathematically, this problem is formulated as:
\begin{equation}
    \begin{aligned}
    \acttraj^{\star}_t &= \text{argmax}_{\acttraj_t \sim \pi(\obs_t, \lang)} \mathbb{E}_{\obstraj_t \sim \dyn(\obs_t, \acttraj_t)} \Big[R\big(\obstraj_t; \lang \big) \Big], 
\end{aligned}
\label{eq:visual_mpc}
\end{equation}
where $\dyn(\obs_t, \acttraj_t)$ is a visual dynamics model that predicts a sequence of future observations ($\obstraj_{t}:=\obs_{t:t+\thor}$) conditioned on an action plan $\acttraj_t$ and the current observation $\obs_t$, and $R$ is a reward function that evaluates the predicted outcomes given the task description ($\lang$). 
This framework is particularly appealing because it can leverage pre-trained generative imitation-based policies ($\pi$) as action plan samplers and align the robot’s behavior with deployment-time task context and preferences without fine-tuning or modifying the base policy.

\textbf{World model as visual dynamics for action plan verification and selection}.
World models enable robots to ``imagine” future observations given current observations and planned actions.
Typically, a world model consists of three key components: an observation encoder model $z_t = \mathcal{E}_{\phi}(o_t)$ that maps a visual observation $o_t$ into a latent state $z_t$, a forward dynamics model $z_{t+1} = \tilde{f}_{\phi}(z_t, a_t)$ that predicts the next latent state conditioned on the current latent state and action, and an observation decoder model $o_t = \mathcal{Q}_{\phi}(z_t)$ that decodes a latent state to visual observation. 
%Different from prior work that assumes access to a simulator that continuously generates world model training data, we consider a world model trained using an \textit{offline} dataset of robot–environment interactions $\mathcal{D}_{WM} = \{(o_t,a_t)_{t=1}^{T}\}_{i=1}^{N}$. 
%In this work, we leverage a world model as the visual dynamics model to forward simulate the future induced by the robot's action plans.
World models have been increasingly adopted as generalized
dynamics models to produce action-conditioned future outcomes for downstream robot policy learning \cite{wu2023daydreamer, zhou2024dino} or verification under the model predictive control framework described in \eqref{eq:visual_mpc} \cite{gaoflip,qi2025strengthening,wu2025foresight}. 

\textbf{Challenge: novel visual distractors cause hallucinations and compromise planning}.
While world models enable robots to imagine future visual outcomes and inform policy learning, the accuracy and confidence about their imaginations heavily depend on the consistency between the training environment and the deployment environment.
The existing effort on exploring world models for robotics applications mostly focuses on training-testing consistent conditions \cite{gaoflip,qi2025strengthening,wu2025foresight}; however, real-world deployment inevitably involves open-world environments where novel visual distractors (such as objects and background elements rarely seen during training) are inevitable.
Current world models are brittle under such conditions: these novel distractors can corrupt the model, leading to hallucinated imaginations that ultimately compromise downstream action plan verification and selection (refer to the motivating example in Figure~\ref{fig:front-figure}).

%% file: sections/method.tex
\section{World Model Reimagination with Test-time Observation Intervention for Visual Model Predictive Control}
\vspace{-0.1cm}
\label{sec:method}

We propose Reimagination with Observation Intervention (\textbf{ReOI}), a \textbf{\textit{test-time}} and \textbf{\textit{plug-in}} strategy that mitigates the impact of novel visual distractors on world model-based policy verification and selection, where a verifier needs to select desired action plans based on visual outcome predictions from a world model. 

Since mitigating dynamic distractors remains a largely open and underexplored challenge, in this work, we focus on mitigating the impact of \textit{static} novel visual distractors on world model–based robot planning. We define visual distractors as static objects and background elements that are not directly related to the task from the task specification. Different from the definition in \cite{hancock2024run}, we do not assume that distractors do not affect the robot's motion to reach the goal.
Our key idea is to first identify and inpaint problematic novel distractors from the current observation to bring it closer to the training distribution, then reimagine future outcomes using the modified input, and finally reintroduce the distractors post-hoc to preserve visual consistency for downstream planning and verification.

% shifting them closer to the model’s training distribution to enable more stable and accurate rollout predictions of robot behavior. 

% After rollout, the unfamiliar distractors are reintroduced at the pixel level to maintain visual consistency for downstream verification. 

% distractors by analyzing physically implausible changes in predicted rollouts,  and then reimagines future outcomes before reintroducing distractors post-hoc to preserve visual consistency for downstream verification.

%
% While this strategy may result in physically unrealistic interactions—such as the robot appearing to pass through reintroduced distractors—it does not compromise policy steering, as the robot needs to avoid any physical contact with such distractors to ensure safe behavior.

\subsection{Observation Intervention Strategy}

\begin{figure*}[t]
  \centering
\includegraphics[width=0.8\textwidth]{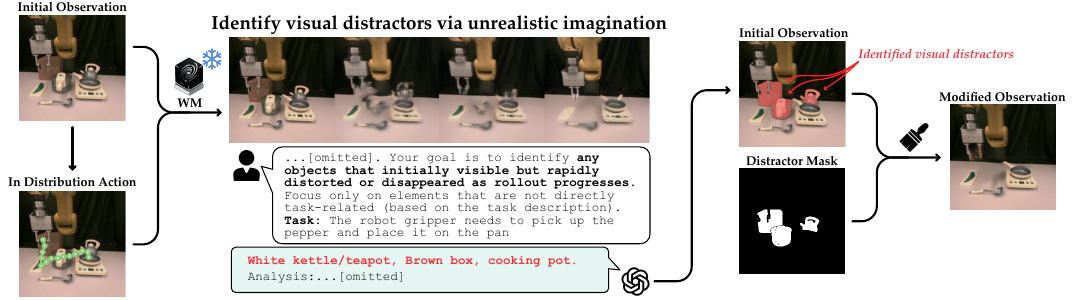}
  \caption{\textbf{Novel visual distractor identification and observation intervention}. We leverage a VLM to analyze the temporal evolution of objects and flag those that undergo rapid distortion across frames as potential novel visual distractors (left-side; more examples and the full prompt can be found in Appendix~\ref{app:distractor-identification}). These problematic distractors are inpainted to bring the observation closer to the world model’s training distribution (right-side).}
  \vspace{-0.35cm}
  \label{fig:vlm-distractor-identification-inpaint}
\end{figure*}

\textbf{Identify novel visual distractors from world model predictions}.
Since the original training data is typically unavailable at deployment time, we identify novel visual distractors using only the world model’s predicted observations.
Our key insight is that visual distractors underrepresented in the training distribution tend to exhibit physically implausible behavior in world model rollouts: although initially visible, they quickly become distorted, disappear, or warp unnaturally as the rollout progresses, even when the input actions are in the distribution (shown in Figure~\ref{fig:vlm-distractor-identification-inpaint}).
This phenomenon arises because the latent dynamics model prioritizes consistent and predictable visual features learned from its training distribution. When a novel distractor appears, the model lacks an accurate latent representation of its dynamics, resulting in initial prediction errors. As predictions unfold autoregressively, these errors compound, causing the distractor's latent representation to degrade progressively. Consequently, pixels corresponding to unfamiliar distractors fail to be reliably propagated through the model's latent dynamics, leading to rapid visual distortion.

Building on this insight, we leverage a VLM's (GPT-4o) visual reasoning ability to identify problematic visual distractors through visual reasoning.
Given the current observation, the world model first rolls out an in-distribution “safety-check” action plan to generate a sequence of predicted observations. The VLM is then prompted to analyze this predicted sequence. Leveraging its open-world visual reasoning capabilities, the VLM examines the temporal evolution of objects and flags those that undergo rapid distortion across frames as potential novel visual distractors. The output from the VLM is a string of object proposals deemed as novel visual distractors (shown in Figure~\ref{fig:vlm-distractor-identification-inpaint}). 

\textbf{Segmentation and inpainting}. Once distractors are identified, we use a segmentation model \cite{ren2024grounded} to localize and segment the corresponding regions.
These regions are then passed to an image inpainting model \cite{chen2024rovi}, which removes the distractors and fills in the masked areas to produce a modified observation (shown in Figure~\ref{fig:vlm-distractor-identification-inpaint}).
This intervention yields an input that more closely aligns with the world model’s training distribution, alleviating hallucinations caused by artifacts from novel distractors.

\subsection{World Model Reimagination with Modified Observation}

After intervention, we use the original world model again to predict the action plan outcomes given the modified input observation. 
While this mitigates hallucinated robot behaviors caused by spurious distractor artifacts, the resulting predictions no longer model the novel visual distractors since they were inpainted.
To preserve visual consistency for downstream robot policy verification, we reintroduce the removed distractors into each predicted observation post-hoc.

\input{sections/figure_patch_distractors}
Specifically, we first decompose each predicted observation into a set of object layers using Grounded-SAM2 \cite{ren2024grounded}, separating elements such as the robot, task-relevant objects, and background. For each layer, we estimate its depth from the predicted frame \cite{oquab2023dinov2}. We also extract the inpainted distractor layer from the original observation and estimate its depth based on its original spatial placement. This distractor layer is then added back into the set of layers for every predicted frame.
Finally, we reconstruct each predicted frame by compositing the layers in back-to-front depth order, ensuring that all objects are rendered with correct occlusion. This layer-wise rendering approach maintains visual realism for downstream planning, even though the distractors were absent from the rollout itself (illustrated in Figure~\ref{fig:reimagination_visual}).

We note that this approach may introduce physically unrealistic interactions, such as the robot gripper may appear to pass through reinserted distractors if they block the robot's motion. However, this does not compromise policy verification. Since the robot must avoid any physical contact with such distractors to ensure safety, any trajectory exhibiting these violations is automatically rejected during the verification process.

\subsection{Action Plan Verification and Selection for Robot Visual Planning}

We deploy ReOI as the visual dynamics model in the context of robot action plan verification, where a verifier needs to select action plans based on visual outcome predictions from ReOI.
While our framework is agnostic to the choice of verifier, we instantiate it with a VLM that assesses each reimagined rollout in the context of the task instruction $\lang$.
The VLM is prompted to identify and reject plans that pose safety concerns (e.g., collisions with novel distractors or non-target objects) and to select the outcome that best aligns with the user’s intent.
If none of the proposed plans are deemed safe or suitable, the VLM can optionally escalate by requesting human intervention.

%% file: sections/figure_patch_distractors.tex
\begin{figure}
\centering
\includegraphics[width=0.9\linewidth]{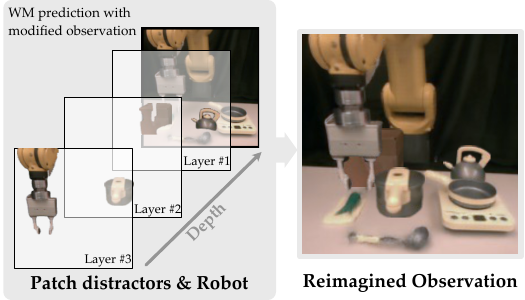}%
\caption{\textbf{Reimagination with modified observation}. Predicted future observations from the modified input are post-processed by reintroducing previously inpainted distractors using depth-aware compositing. This ensures correct occlusion, e.g., the package box appears \textit{behind} the gripper.} 
\vspace{-0.5cm}
\label{fig:reimagination_visual}
\end{figure}

%% file: sections/experiment_setup.tex
\section{Experiment Setup}
\vspace{-0.2cm}

\textbf{Testing environment.}
We conduct our evaluations using a Fanuc LR Mate 200iD/7L 6-DoF robot in a real-world robotic manipulation setup. The robot is tasked with picking and placing objects in a toy kitchen environment.

\textbf{Pre-trained imitation-based generative policy.} We use a Diffusion Policy \cite{chi2023diffusionpolicy} as the imitation-based action plan sampler. The model takes the current image observations from the wrist and third-person cameras as inputs to predict a distribution of the robot’s future action plans (each action is a collection of a 3D waypoint and a gripper control signal). We use $120$ multi-mode teleoperated demonstrations to train the policy.

\textbf{World model training.} We use the \dinowm \cite{zhou2024dino} as our base world model. DINO-WM leverages pre-trained DINOv2 \cite{oquab2023dinov2} representation to encode visual observations and predict action outcomes directly in the DINOv2 latent representation space. 
We train the world model using 500 robot–environment interaction trajectories, with 200 trajectories sampled by rolling out the pre-trained diffusion policy and 300 random exploration trajectories. We separately fine-tune a DINOv2 feature decoder using images from the testing environment to map the predicted latent representations back to visual observations. More details can be found in Appendix~\ref{app:world-model-training}.

%% file: sections/results.tex
\section{Results}
\vspace{-0.1cm}
 \subsection{On the Effect of Novel Visual Distractors on World Model Predictions}

% In this section, we qualitatively and quantitatively evaluate the impact of novel visual distractors on world model predictions.

\begin{figure*}[t]
  %\vspace{-0.8cm}
  \centering
\includegraphics[width=0.8\textwidth]{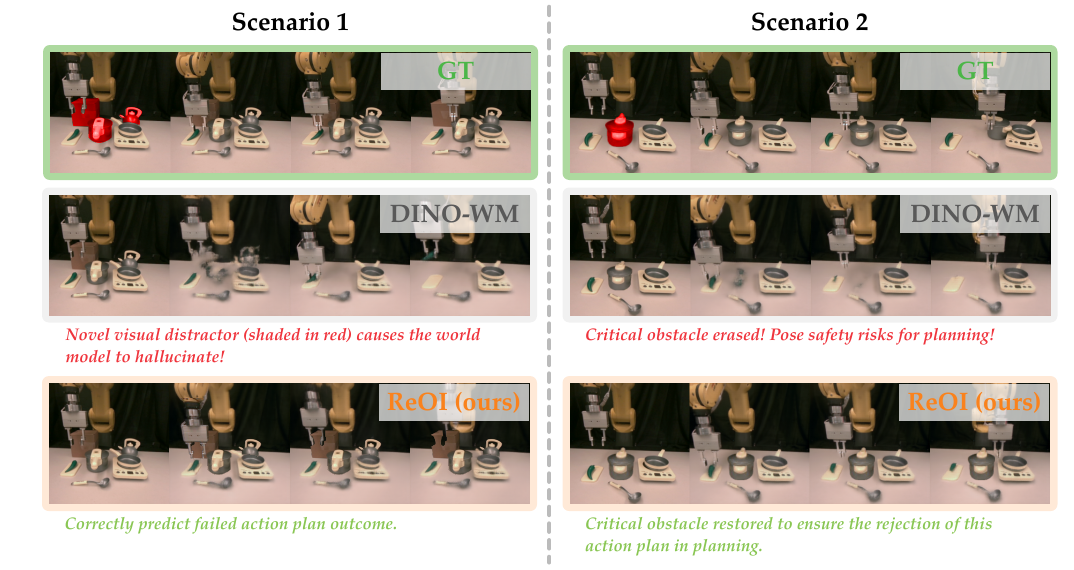}
  \caption{\textbf{Qualitative examples of prediction quality}. The presence of unfamiliar distractor objects causes DINO-WM to hallucinate and predict that a failed action plan would succeed (left side, middle row) and mistakenly erase the target object (the green pepper) from the predicted observations (right side, middle row). \ours (bottom row) effectively mitigates the effect of the novel visual distractors and produces visual outcomes more aligned with the ground truth (top row).}
  \vspace{-0.25cm}
  \label{fig:wm-pred-qualitative-examples}
\end{figure*}

\textbf{Qualitative example}. 
In Figure \ref{fig:wm-pred-qualitative-examples}, we demonstrate two representative examples illustrating how novel visual distractors (highlighted in red in the first frame of each ground-truth rollout) degrade the predictive performance of the world model.
Across these examples, the presence of unfamiliar distractors causes \dinowm to hallucinate and predict that a failed action plan (one that would not successfully pick up the green pepper) would succeed (left) and mistakenly erase the target object (the green pepper) from the predicted observations (right).
In contrast, \ours produces significantly more accurate and visually consistent predictions aligned with the true task execution by reimagining future observations through test-time observation intervention.

\begin{table}[h]
\centering
\begin{tabular}{l|c c | c c}
\toprule
& \multicolumn{2}{c|}{\textbf{Full obs. eval.}} & \multicolumn{2}{c}{\textbf{In-distribution component eval.}} \\
\textbf{} & {SSIM $\uparrow$} & {LPIPS $\downarrow$} & {SSIM} $\uparrow$ &  {LPIPS$\downarrow$} \\
\toprule
\textbf{\dinowm}  & 0.51 & 0.17 & 0.63 & 0.14 \\ 
\textbf{\ours} & \cellcolor[HTML]{c4f2cb}\textbf{0.94} & \cellcolor[HTML]{c4f2cb}\textbf{0.04} &  \cellcolor[HTML]{c4f2cb}\textbf{0.79} & \cellcolor[HTML]{c4f2cb}\textbf{0.09}\\
\toprule
\end{tabular}

%\vspace{-0.2cm}
\caption{\textbf{Quantitative evaluations of a world model’s predicted visual action outcomes}. We measure SSIM and LPIPS to evaluate the structural and perceptual similarity of the predicted visual action outcome. 
The left side shows results on full predicted observations, while the right side focuses on task-relevant, in-distribution objects after inpainting distractors from both the predictions and ground truth.
\ours enables the world model to produce more accurate visual predictions and effectively mitigates the hallucination of in-distribution objects caused by novel visual distractors.}
\vspace{-0.3cm}
\label{tab:wm_pred_quality}
\end{table}

\textbf{Quantitative evaluation: overall prediction quality}. 
We use two standard metrics, SSIM (Structural Similarity Index) \cite{wang2004image} and LPIPS (Learned Perceptual Image Patch Similarity) \cite{zhang2018unreasonable}, to evaluate the quality of the world model’s predicted visual action outcomes. 
SSIM measures structural similarity based on contrast and spatial consistency, with higher scores indicating closer alignment. LPIPS evaluates perceptual similarity using deep feature comparisons, with lower scores indicating greater visual similarity.
As shown in Table~\ref{tab:wm_pred_quality}, \ours achieves better SSIM and LPIPS scores compared to \dinowm. These improvements indicate that by shifting the input observation closer to the training distribution through test-time intervention and reimagining futures from this intervened input, our approach enables the world model to produce more accurate and robust predictions.

\textbf{Quantitative evaluation: ReOI effectively mitigates the hallucination of in-distribution objects caused by novel visual distractors.}
Since \ours reimagines at test time and explicitly reintroduces the novel visual distractors that  \dinowm struggles to predict, the previous evaluation does not fully disentangle the effect of these distractors on the world model’s ability to predict the dynamics of in-distribution components such as the robot and target objects.
To ablate this effect, we further inpaint the identified distractors in both the ground truth and predicted observations and then measure SSIM and LPIPS over the inpainted (distractor-free) predictions. The right side of Table~\ref{tab:wm_pred_quality} shows that \ours still achieves better consistency and perceptual similarity compared to \dinowm.
These results indicate that without test-time intervention, novel visual distractors not only degrade the overall visual quality (e.g., distractors distorted across frames) but also corrupt the predicted dynamics of in-distribution objects (robot, target object), leading to hallucinated motions.

\subsection{On the Value of Test-time Observation Intervention for Visual Planning}

\begin{figure*}[t]
  %\vspace{-0.8cm}
  \centering
  \includegraphics[width=0.8\textwidth]{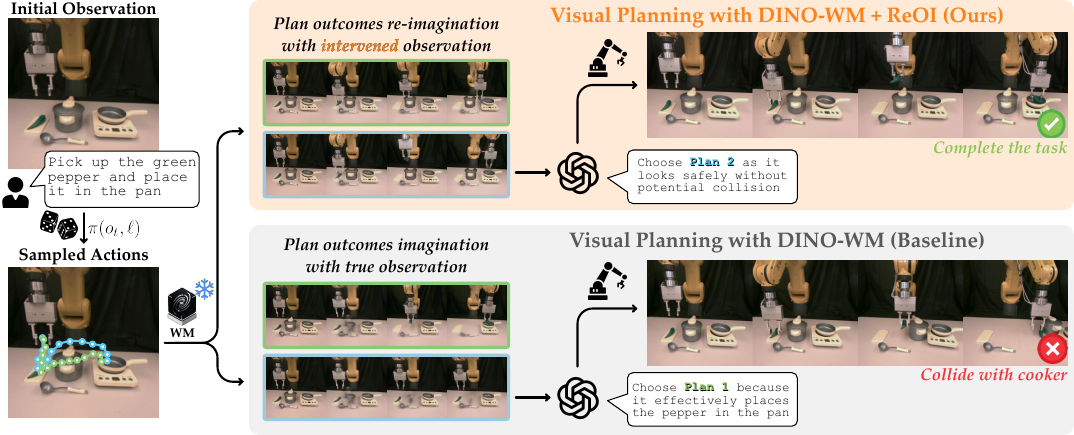}
  \caption{\textbf{Qualitative example of robot visual planning}. 
  The novel distractor object causes \dinowm to hallucinate and erase a critical obstacle, leading the VLM plan verifier to incorrectly accept an unsafe action plan.
  In contrast, \ours generates visual outcome predictions that better aligned with the reality, allowing the plan verifier to recognize the potential collision with the distractor and correctly select a safe action plan for execution.}
  \vspace{-0.6cm}
  \label{fig:system-demo-visual-example-1}
\end{figure*}

In this section, we evaluate the system-level performance in the context of action plan verification, where the VLM-based verifier needs to select desired action plans based on visual outcome predictions from the world model. More detailed component-level evaluation (visual distractor identification accuracy and plan verification accuracy) can be found in Appendix~\ref{app:system-eval}).

\textbf{Baselines.}
Our approach mitigates the impact of novel environmental variations on downstream planning by applying test-time observation intervention and reimagination.
Alternatively, another intervention strategy is to reject untrustworthy world model action outcome predictions at test time to ensure safety.
In addition to comparing against the base \dinowm, we compare our approach against \trust \cite{knuth2021planning}, which finds a region (a union of $r$-balls about the subset of the world model training data) where the learned visual dynamics model is deemed reliable for downstream planning and rejects world model predictions if the input observation and action plan pair falls outside of the trust region. 
%TrustRegion leverages an estimated Lipschitz constant (bound how much outputs change with respect to a change in the inputs) of the world model visual outcome prediction error to obtain an error bound for a test-time input observation. By defining a maximum acceptable visual dynamics prediction error threshold, we could find the corresponding trust region such that any input observation and action plan pair within this region produces an error bounded by the threshold.
More details about this baseline can be found in Appendix~\ref{app:trust-region}.

\textbf{Metric.} We measure the task success rate and collision rate to evaluate the system-level performance of the robot visual planning. For each method, we conduct 10 trials with randomly initialized task configurations and report the average success rate. A trial is considered successful if the robot successfully and safely completes the task.

\begin{table}[h]
\centering
\begin{tabular}{l|c c}
\toprule
\textbf{} & {\textbf{Success Rate} $\uparrow$} & {\textbf{Collision Rate} $\downarrow$}\\
\toprule
\dinowm & 0.20 & 0.40\\
\trust & 0.00 & \cellcolor[HTML]{c4f2cb}\textbf{0.10}\\
\ours & \cellcolor[HTML]{c4f2cb}\textbf{0.70} & \cellcolor[HTML]{c4f2cb}\textbf{0.10}\\
\toprule
\end{tabular}
\vspace{-0.2cm}
\caption{\textbf{Task Success Rate}. \ours enables more effective and safe visual planning compared to baselines.}
\vspace{-0.2cm}
\label{tab:system-level-eval}
\end{table}

\textbf{System-level result.} We present the system-level quantitative result in Table~\ref{tab:system-level-eval}. The results show that \ours achieves a significantly higher task success rate compared to both \dinowm and \trust.
Notably, \ours maintains a low collision rate comparable to the conservative \trust, while being substantially safer than \dinowm that directly uses the observations for future prediction.
These results suggest that \ours enables more effective and safe visual planning by modifying the input observation and reimagining action outcomes at test time. We show two qualitative examples in Figure~\ref{fig:front-figure} and Figure~\ref{fig:system-demo-visual-example-1} to demonstrate the effectiveness of our approach.

%% file: sections/conclusion.tex
\section{Conclusion}

World models enable robots to ``imagine” future observations given current observations and planned actions, and have been increasingly adopted as generalized dynamics models to facilitate robot learning.
Despite their promise, these models remain brittle when encountering novel visual distractors. Specifically, novel distractors can corrupt action outcome predictions, causing downstream failures when robots rely on the world model imaginations for planning or action verification. 
In this work, we proposed Reimagination with Observation Intervention, a simple yet effective test-time strategy that enables world models to predict more reliable action outcomes in open-world scenarios where novel and unanticipated visual distractors are inevitable.
We validated our approach on robotic manipulation tasks in the context of action verification, where the verifier needs to select desired action plans based on predictions from a world model.
Results showed that our approach is robust to both in-distribution and unfamiliar visual distractors. Notably, it improved task success rates by up to \textbf{3×} in the presence of novel distractors, significantly outperforming action verification that relies on world model predictions without imagination interventions.

%% file: sections/appendix.tex
\section{Novel Distractor Identification}
\label{app:distractor-identification}
Our key insight is that visual distractors underrepresented in the training data often behave implausibly during world model rollouts—quickly distorting, disappearing, or warping unnaturally over time. To detect such cases, we prompt the VLM with two frames from the world model’s predicted rollout: the initial frame, which contains the full scene context including potential distractors, and the fifth frame, where empirical analysis shows that distractors are most likely to have degraded or disappeared. This selection is based on our observation that implausible distractor behavior typically manifests within the first four steps of rollout prediction. To enable precise spatial reasoning, we generate segmentation masks for the initial frame using Grounded-SAM2~\cite{ren2024grounded} and overlay unique mask IDs directly onto the segmentation map. The VLM is then presented with the ID-labeled segmentation image alongside the initial and fifth frames, and is prompted to identify which regions (i.e., mask IDs) correspond to objects that have disappeared or changed unnaturally over time.

\begin{figure*}[htbp]
    \centering
    \includegraphics[width=0.8\linewidth]{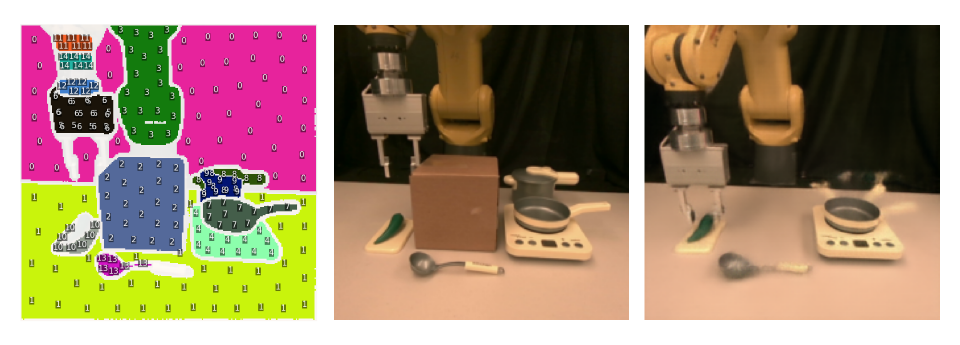}
    \caption{The VLM is prompted to identify visual distractors by reasoning about missing objects between the initial frame (middle) and the fifth frame (right) of the world model rollout, using mask IDs from the semantic segmentation (left).}
    \label{fig:gpt_input}
\end{figure*}

\begin{tcolorbox}[
rounded corners,
title=Prompt for novel distractor identification
]
You are a helpful assistant for a robotic system that analyzes images to detect changes and disappearing objects.

I'll show you three images:\\
1. img\_init\_dis - The initial image\\
2. img\_wm - The same scene after some changes\\
3. mask\_overlay - The initial image with numbered masks/patches created by a segmentation algorithm\\
\\
Your task is to identify which numbered patches from the mask\_overlay image exist in img\_init\_dis but have disappeared in img\_wm. Look carefully at each numbered patch and determine if the corresponding object still present in img\_wm.\\
\\
Hint: This is a kitchen environment with a robot arm, try only detect patches that the corresponding object are on the white table.\\
\\
Please provide ONLY a list of the missing patch numbers with no additional explanation. For example, if patches 2, 5, and 7 have disappeared, just respond with: [2, 5, 7].\\
\end{tcolorbox}

\begin{figure*}[htbp]
    \centering
    \includegraphics[width=0.8\linewidth]{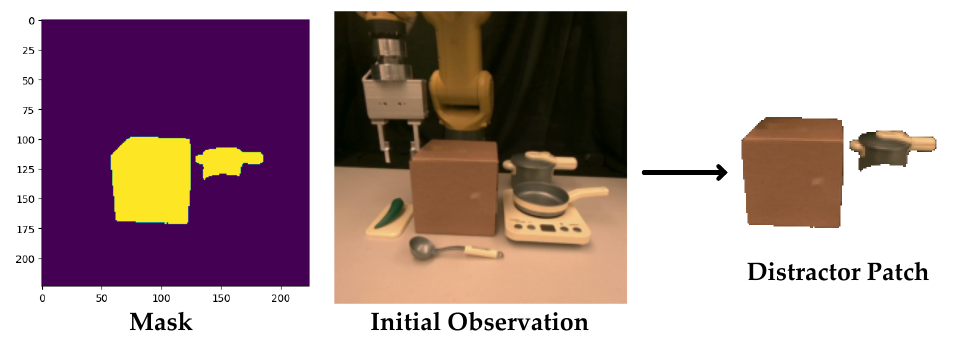}
    \caption{Distractor patches extracted from the initial observation based on the identified mask.}
    \label{fig:distractor_patch}
\end{figure*}

Once the VLM produces its response—typically in the form of a list of missing or degraded patch IDs (e.g., \texttt{Missing patches:[2, 8, 9]})—we use the identified mask indices to extract the corresponding image patches from the initial observation. These patches represent regions that are likely to be visual distractors causing inconsistencies in the rollout. Figure~\ref{fig:distractor_patch} shows an example of these extracted distractor patches.

\section{World Model Training}
\label{app:world-model-training}

\subsection{\dinowm Training}
We adopt \dinowm \cite{zhou2024dino} as our base world model. The hyperparameters used for training are listed in Table~\ref{tab:wm_param}. Training is performed on 500 robot–environment interaction trajectories, comprising 200 rollouts from a pre-trained diffusion policy and 300 from random exploration. Each trajectory spans 24 steps with a 0.75-second interval between steps in the real world. We omit frameskip during training, as the existing time gap is sufficient—predicting further into the future would impair model accuracy. All models are trained on a single Nvidia A6000.

\begin{table}[htbp]
    \centering
    \caption{Hyperparameters for \dinowm training.}
    \begin{tabular}{lc}
    \toprule    
        Name & Value \\
    \midrule
        seed & 0 \\
        epochs & 300 \\
        batch\_size & 64 \\
        save\_every\_x\_epoch & 10 \\
        reconstruct\_every\_x\_batch & 500 \\
        num\_reconstruct\_samples & 6 \\
        encoder\_lr & 1e-6 \\
        decoder\_lr & 2e-5 \\
        predictor\_lr & 5e-4 \\
        action\_encoder\_lr & 5e-4 \\
        img\_size & 224 \\
        frameskip & 1 \\
        concat\_dim & 1 \\
        saved\_folder & null \\
        normalize\_action & True \\
        action\_emb\_dim & 10 \\
        num\_action\_repeat & 1 \\
        proprio\_emb\_dim & 10 \\
        num\_proprio\_repeat & 1 \\
        num\_hist & 3 \\
        num\_pred & 1 \\
        has\_predictor & True \\
        has\_decoder & True \\
    \bottomrule
    \end{tabular}
    \label{tab:wm_param}
\end{table}

\begin{figure*}
    \centering
    \includegraphics[width=0.9\linewidth]{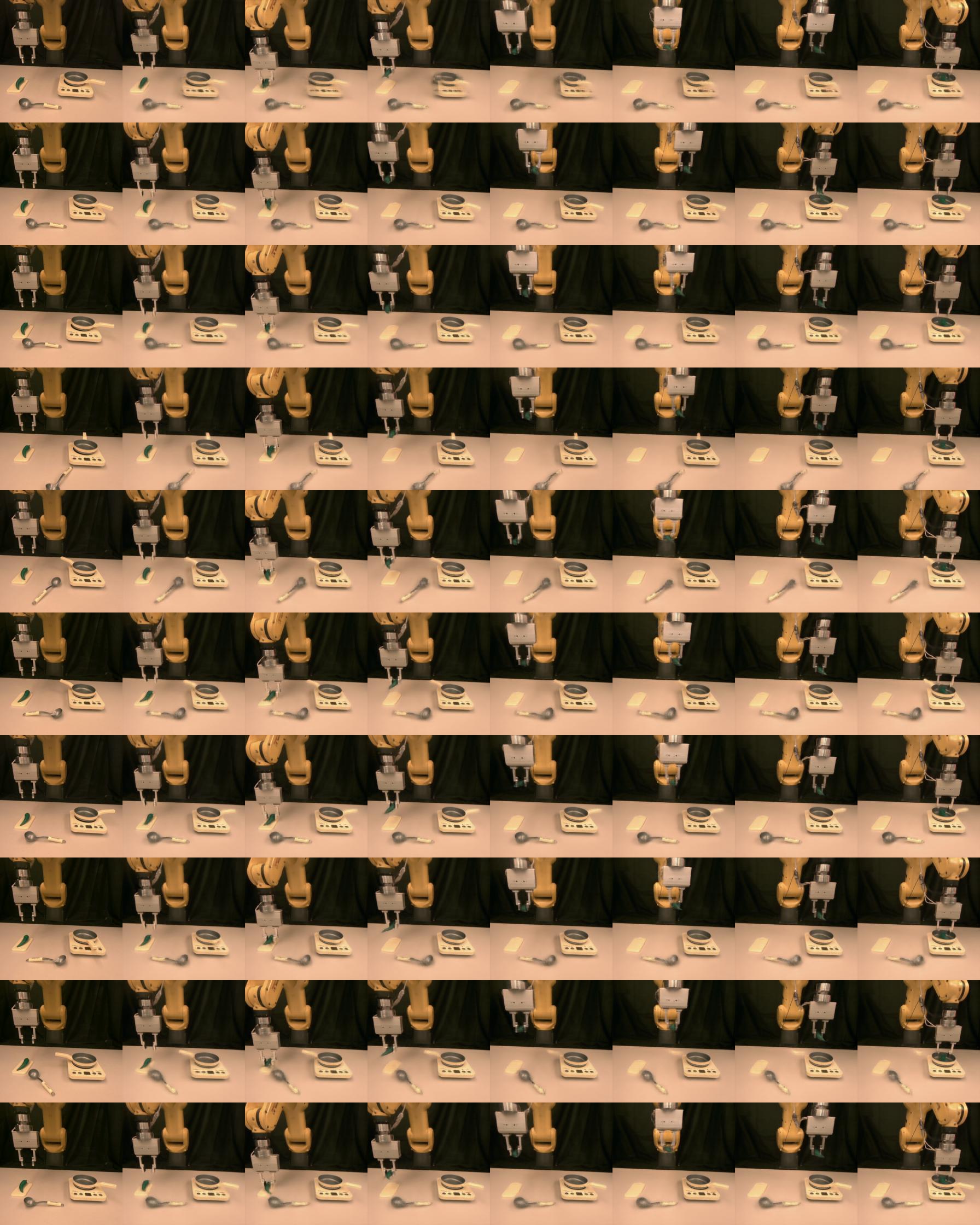}
    \caption{Trajectories predicted with \dinowm with 10 different initial observations without visual distractors.}
    \label{fig:wm_rollout_id}
\end{figure*}

All 500 training trajectories are collected in a controlled kitchen environment, where the robot is tasked with transporting a green pepper from a cutting board—randomly positioned on the left side of the table—to a cooking pan placed on an induction stove with varying orientations. To introduce additional variability, a spoon is also randomly placed on the table surface during data collection. As illustrated in Figure~\ref{fig:wm_rollout_id}, \dinowm is able to accurately predict future states with fine-grained visual detail, maintaining both temporal and spatial consistency across frames. It also demonstrates robustness to environmental randomization within the training domain.

\begin{figure*}
    \centering
    \includegraphics[width=0.9\linewidth]{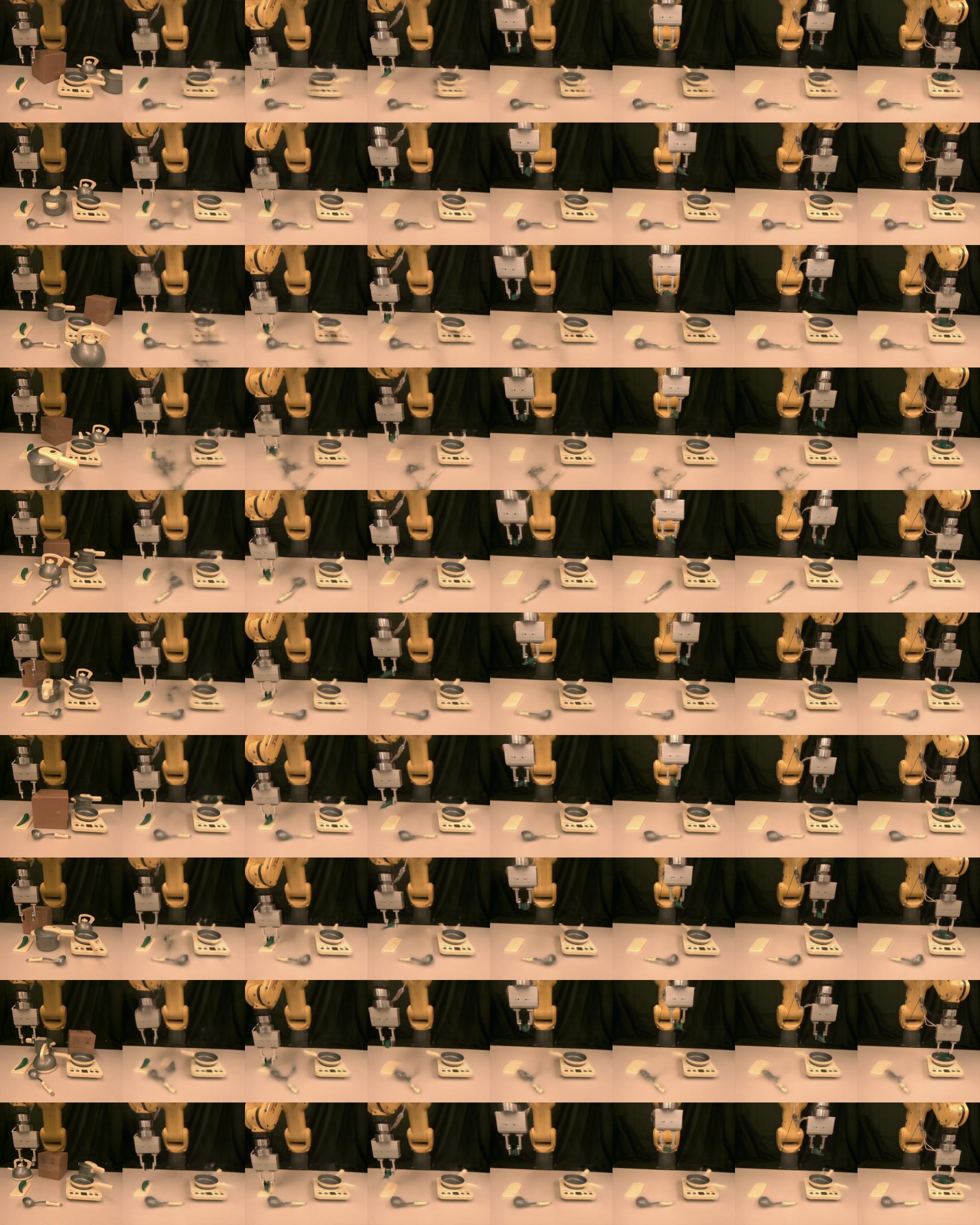}
    \caption{Trajectories predicted with \dinowm with 10 different initial observations with visual distractors.}
    \label{fig:wm_rollout_ood}
\end{figure*}

However, when we introduce novel visual distractors—such as an Amazon box, a kettle, and a high-pressure pot—randomly positioned within the scene, the predictive performance of \dinowm significantly deteriorates. In these out-of-distribution scenarios, parts of the predicted images become visibly distorted, disappear altogether, or warp in physically implausible ways across the rollout sequence, as shown in Figure~\ref{fig:wm_rollout_ood}.

\begin{figure*}
    \centering
    \includegraphics[width=0.9\linewidth]{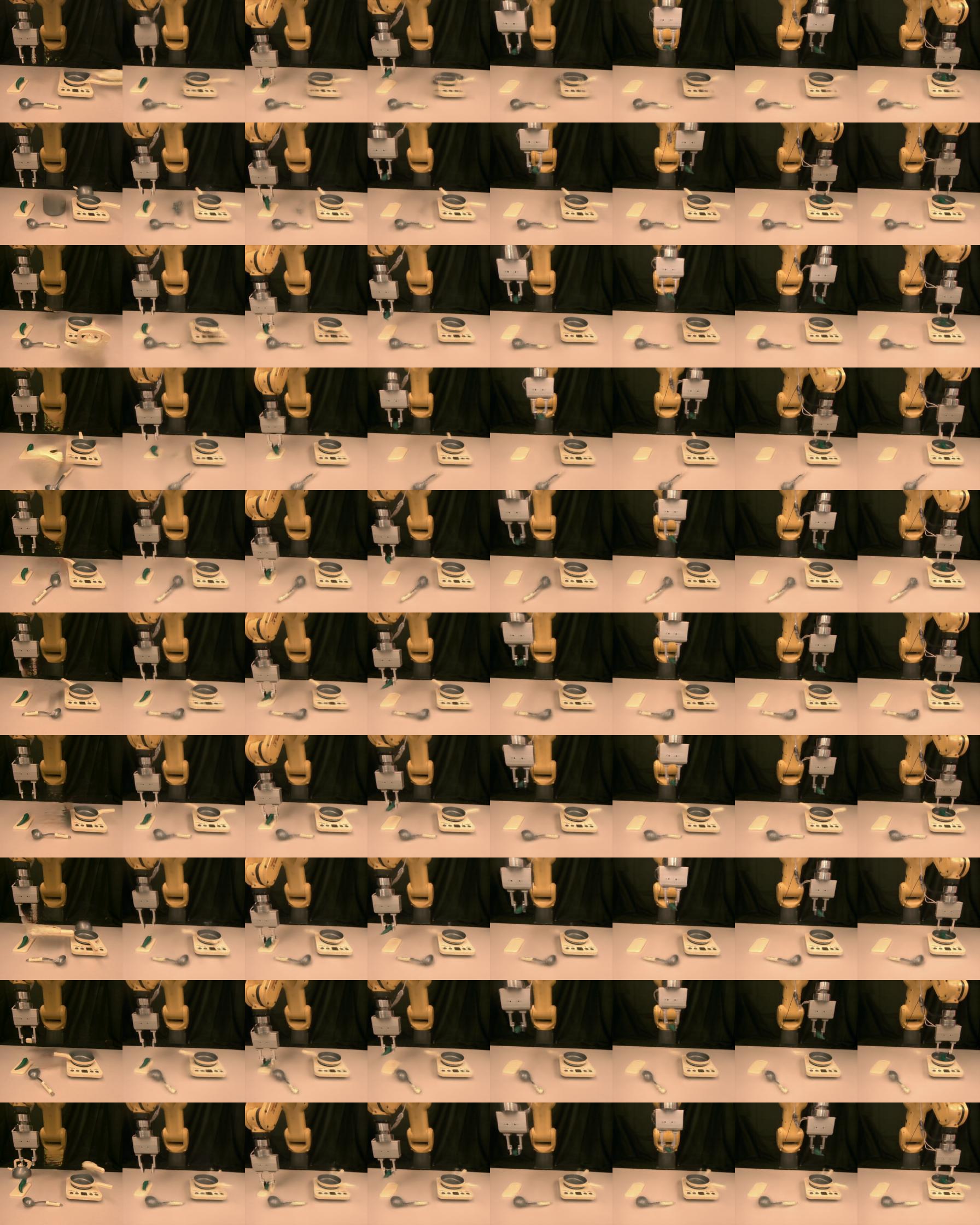}
    \caption{Trajectories reimaginated by \dinowm with 10 modified initial observations.}
    \label{fig:wm_rollout_recon}
\end{figure*}

Figure~\ref{fig:wm_rollout_recon} presents the reimagined rollout results using modified input observations for the same 10 initial states containing visual distractors. While some artifacts remain—primarily due to limitations in the segmentation and inpainting models—the severe hallucinations observed in the original rollouts have been effectively mitigated. Notably, the differences in row 4 and row 9 between Figure~\ref{fig:wm_rollout_ood} and Figure~\ref{fig:wm_rollout_recon} highlight the improved visual stability and consistency achieved through our intervention.

\subsection{DINOv2 Feature Decoder}
To enable accurate reconstruction of visual observations from predicted latent representations in the testing environment, we perform a dedicated fine-tuning of the DINOv2 feature decoder. This process is carried out separately from the original world model training. We first collect a new dataset consisting of 80 trajectories captured within the testing environment, which intentionally includes various visual distractors not present during the initial training phase. These trajectories are used to better adapt the feature decoder to the distributional shift introduced by the novel visual elements.

We then initialize the training pipeline by loading a pre-trained \dinowm checkpoint, which contains the previously trained dynamics model and feature decoder. However, during fine-tuning, we freeze the parameters of the dynamics model entirely to preserve its learned temporal representations and focus solely on updating the parameters of the DINOv2 feature decoder. This ensures that only the visual decoding pathway is adapted to the new environment, without altering the predictive capabilities of the dynamics model itself.

The fine-tuning procedure is conducted over a single epoch using the newly collected dataset. This one-pass update is sufficient to adjust the decoder’s mapping from latent space to image space, improving reconstruction fidelity in the presence of unfamiliar distractors while maintaining consistency with the world model’s learned representations.

\section{Trust Region for Unconfident World Model Prediction Rejection}
\label{app:trust-region}

Our approach mitigates the impact of novel environmental variations on downstream planning by applying test-time observation intervention and reimagination.
Alternatively, another intervention strategy is to reject untrustworthy world model action outcome predictions at test time to ensure safety.
In addition to comparing against the base \dinowm, we compare our approach against \trust \cite{knuth2021planning}, which finds a region where the learned visual dynamics model is deemed reliable for downstream planning and rejects world model predictions if the input observation and action plan pair falls outside of the trust region.

\textbf{Lipschitz constant} of a function quantifies the maximum rate at which the function's output can change with respect to changes in its input. 
Formally, a function $f$ is Lipschitz continuous if there exists a constant $L \geq 0$ such that for all inputs $x_1$ and $x_2$, the following holds:
$\|f(x_1) - f(x_2)\| \leq L \cdot \|x_1 - x_2\|$.
Here, $L$ is denotes the Lipschitz constant, and it provides an upper bound on the function’s sensitivity.

TrustRegion leverages an estimated Lipschitz constant (bound how much outputs change with respect to a change in the inputs) of the world model visual outcome prediction error to obtain an error bound for a test-time input observation. By defining a maximum acceptable visual dynamics prediction error threshold, we could find the corresponding trust region such that any input observation and action plan pair within this region produces an error bounded by the threshold.

Let $T$ denote a trust region (a union of $r$-balls about the subset of the world model training data), $b_T$ denote the dispersion of the region, and $e_T$ be the maximum training error of the trained world model. Then for any input (a pair of an initial observation and an action plan) within the $T$, its world model prediction error is bound by
\begin{equation}
    \epsilon = L_{T} b_{T} + e_{T}.
\end{equation}

\textbf{Implementation.} We estimate the Lipschitz constant of the world model's visual action plan outcome prediction error. In this work, we use the L2 difference between the world model's predicted latent state trajectory and the ground-truth latent state (DinoV2 embedding) trajectory given the same initial observation and the action plan as the prediction error.
Following the procedure from \cite{knuth2021planning}, we begin by selecting training data points whose prediction error is below a threshold of $750$ (an empirically chosen value below which the model’s predictions typically highly align well with ground truth). These filtered data points are used to initialize the trust region $T_0$, defined as a union of $r_0$-balls centered around the selected points. In our experiments, we set $r_0$ as a unit ball with a radius of $0.1$, and the corresponding Lipschitz constant estimated within $T_0$ is 0.84.
To maximize the flexibility of the trust region for planning, we progressively increase $r$ to expand the region, while ensuring that the Lipschitz constant remains within acceptable bounds ($<1$). The full procedure for this expansion follows Algorithm 2 in \cite{knuth2021planning}. The final trust region has a Lipschitz constant of $0.93$ and the error bound is 1160.

\section{Extended Results for World Model-based Robot Policy Verification}
\label{app:system-eval}

\begin{figure}[htbp]
  \centering
\includegraphics[width=\linewidth]{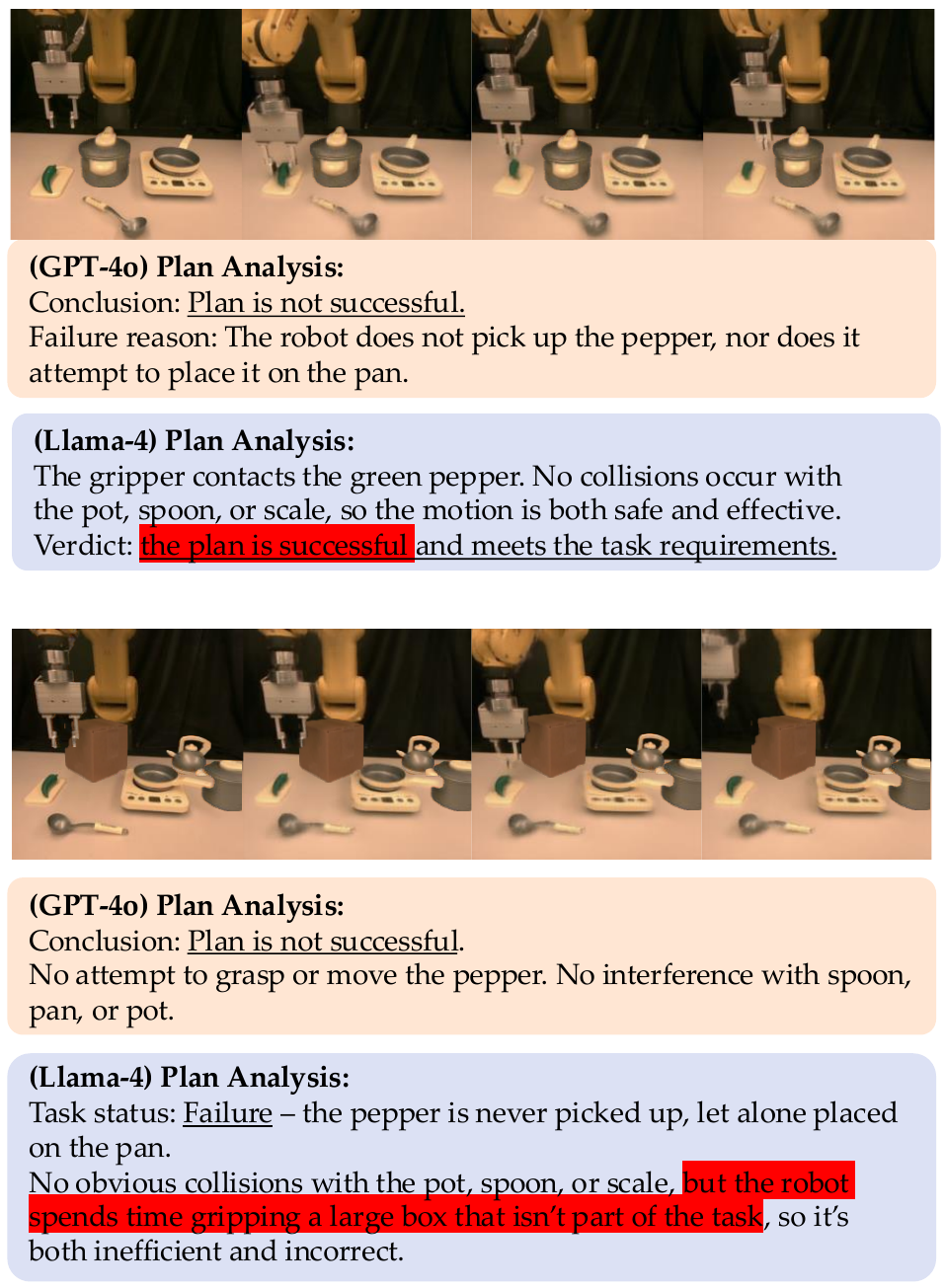}
  \caption{\textbf{Qualitative comparison of VLMs’ plan verification capabilities}.}
  \label{fig:response-compare}
\end{figure}

\subsection{Action Plan Verification and Selection}

We use a VLM as the verifier to evaluate each sampled action plan's visual outcome in the context of the task instruction $\lang$.
The VLM receives a sequence of predicted observations from the world model (sampled as four concatenated frames) and is prompted to identify and reject plans that pose safety concerns—such as collisions with novel distractors or non-target objects—and to select the outcome that best aligns with the user’s intent.
If none of the proposed plans are deemed safe or suitable, the VLM can optionally escalate by requesting human intervention.
\begin{figure*}[t]
  \centering
\includegraphics[width=1.\textwidth]{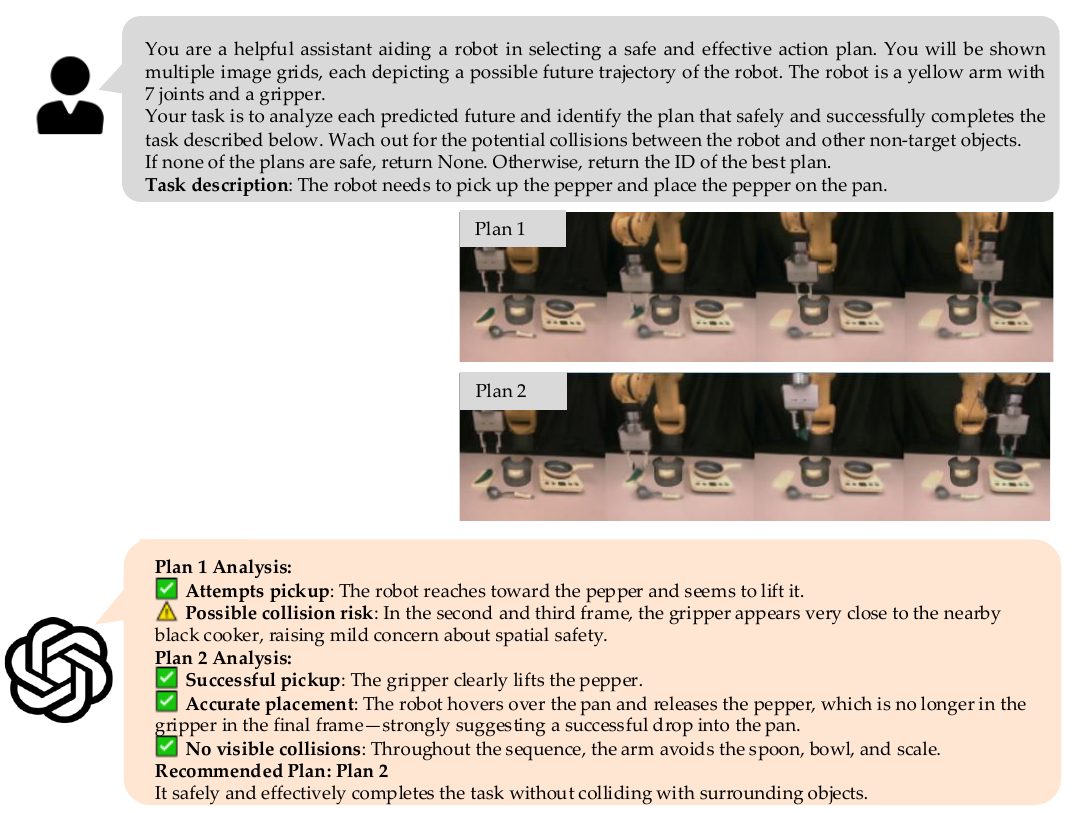}
  \caption{\textbf{Qualitative example of the VLM-based plan verification and selection}. }
  \vspace{-0.2cm}
  \label{fig:prompt-template}
\end{figure*}

\begin{table}[ht]
\centering
\begin{tabular}{l|c c}
\toprule
\textbf{} & {\textbf{GPT-4o}} & {\textbf{LLaMa 4}}\\
\toprule
Verification Accuracy & 0.94 & 0.82\\
Selection Accuracy & 0.85 & 0.64\\
\toprule
\end{tabular}
\caption{\textbf{Plan verification and selection accuracy}.}
\label{tab:verifier_compare}
\end{table}

\textbf{Plan verification and selection accuracy.}
We evaluate two off-the-shelf vision-language models (VLMs), GPT-4o and LLaMA 4, as verifiers without any task-specific fine-tuning.
Table~\ref{tab:verifier_compare} reports two metrics for each VLM: verification accuracy and selection accuracy.
Verification accuracy measures the percentage of trials in which the VLM correctly rejects an unsafe or failed action plan.
Selection accuracy measures the percentage of trials in which the VLM either correctly selects the most desirable action plan or correctly rejects all options if none are safe.
Both metrics are computed based on visual outcome predictions generated by \ours, with ground-truth labels provided by a human expert.
In Figure~\ref{fig:response-compare}, we show response comparisons from the GPT-4o and LLaMa-4.

\textbf{Prompt template}. Figure~\ref{fig:prompt-template} illustrates the system prompt used during the verification stage, along with example responses from the VLM.